\documentclass[sigconf]{acmart}

\usepackage{booktabs} % For formal tables
\usepackage{array}  % ensure this is in your preamble
\usepackage{ragged2e} % only if you still want justified text manually
\usepackage{subfig}
\usepackage{acmart-taps}

\setcopyright{rightsretained}

\acmConference[ICVGIP'25]{16th Indian Conference on Computer Vision, Graphics and Image Processing}{December 2025}{Mandi, India}
\acmYear{2025}
\copyrightyear{2025}

\acmPrice{15.00}

\begin{document}

\title{ChestGPT: Integrating Large Language Models and Vision Transformers for Disease Detection and Localization in Chest X-Rays}
\titlenote{Produces the permission block, and copyright information}

\author{Shehroz S. Khan}
\affiliation{%
  \institution{College of Engineering and Technology, American University of the Middle East}
  \city{Kuwait}
  \country{Egaila, 54200, Kuwait}
}
\email{shehroz.khan@aum.edu.kw}

\author{Petar Przulj}
\affiliation{%
  \institution{Faculty of Applied Science and Engineering, University of Toronto}
  \city{Toronto}
  \state{Ontario}
  \country{Canada}
}
\email{petar.przulj@mail.utoronto.ca}

\author{Ahmed Ashraf}
\affiliation{%
  \institution{Department of Electrical and Computer Engineering, University of Manitoba}
  \city{Winnipeg}
  \state{Manitoba}
  \country{Canada}
}
\email{ahmed.ashraf@umanitoba.ca}

\author{Ali Abedi}
\affiliation{%
  \institution{KITE Research Institute, Toronto Rehabilitation Institute, University Health Network}
  \city{Toronto}
  \state{Ontario}
  \country{Canada}
}
\email{ali.abedi@uhn.ca}

\renewcommand{\shortauthors}{}
\renewcommand{\shorttitle}{ChestGPT: Integrating Large Language Models and Vision Transformers for Disease Detection and \\ Localization in Chest X-Rays}

\begin{abstract}

The global demand for radiologists is increasing rapidly due to a growing reliance on medical imaging services, while the supply of radiologists is not keeping pace. Advances in computer vision and image processing technologies present significant potential to address this gap by enhancing radiologists' capabilities and improving diagnostic accuracy. Large language models (LLMs), particularly generative pre-trained transformers (GPTs), have become the primary approach for understanding and generating textual data. In parallel, vision transformers (ViTs) have proven effective at converting visual data into a format that LLMs can process efficiently. In this paper, we present ChestGPT, a deep-learning framework that integrates the EVA ViT with the Llama 2 LLM to classify diseases and localize regions of interest in chest X-ray images. The ViT converts X-ray images into tokens, which are then fed, together with engineered prompts, into the LLM, enabling joint classification and localization of diseases. This approach incorporates transfer learning techniques to enhance both explainability and performance. The proposed method achieved strong global disease classification performance on the VinDr-CXR dataset, with an F1 score of 0.76, and successfully localized pathologies by generating bounding boxes around the regions of interest. We also outline several task-specific prompts, in addition to general-purpose prompts, for scenarios radiologists might encounter. Overall, this framework offers an assistive tool that can lighten radiologists' workload by providing preliminary findings and regions of interest to facilitate their diagnostic process.

\end{abstract}




\maketitle

\section{Introduction}
The field of radiology is under strain as medical imaging demand increases faster than the supply of available radiologists \cite{cao2023current}. This imbalance results in heavier workloads for radiologists and potential delays in patient diagnoses, which can have serious consequences \cite{limb2022shortages}. To address this issue, recent advancements in computer vision and deep learning offer promising avenues for automated support systems that could enhance diagnostic accuracy \cite{akhter2023ai,hosny2018artificial}.

Large language models (LLMs) have recently gained significant attention due to their ability to process and understand text at an advanced level. Models such as generative pre-trained transformers (GPTs) \cite{radford2018improving} are widely applied across various domains, including healthcare, for tasks such as text generation, question answering, and managing complex information \cite{thirunavukarasu2023large,tang2025rehabilitation}. In parallel, vision transformers (ViTs) \cite{fang2023eva,han2022survey} have shown substantial progress in image analysis. ViTs can process visual data, such as X-rays, and convert it into a format that LLMs can utilize, enabling integrated image-text processing. This capability significantly improves the interpretation and analysis of medical images, particularly in radiology \cite{nerella2024transformers,yan2022radbert}.

Recent studies have explored the integration of LLMs with vision models for medical imaging, especially in the context of chest X-rays \cite{wang2021self,monajatipoor2021berthop,wiehe2022language,akhter2023ai,lee2023unixgen,li2024prompt,thawakar2024xraygpt,cho2024pretraining,keicher2024flexr}. These works have demonstrated that ViTs effectively extract salient features from X-ray images, which LLMs can then leverage to generate diagnostic reports or predict disease outcomes. However, several challenges remain, particularly in accurately localizing pathological regions and improving model interpretability. Moreover, the development of an end-to-end framework that combines both classification and localization within a unified system remains underexplored, indicating a gap in current research.

In this paper, we introduce ChestGPT, a novel deep learning framework that integrates the EVA ViT \cite{fang2023eva} with the LLaMA 2 LLM developed by Meta AI \cite{touvron2023llama}. In this framework, chest X-ray images are first transformed into visual tokens using the ViT, which are then processed by the LLM along with carefully engineered prompts. ChestGPT performs both disease classification and localization, identifying the presence of conditions and highlighting their locations within the image. The framework further employs prompt engineering techniques to enhance interpretability and performance, making it a viable tool for clinical use. We evaluate ChestGPT on the VinDr-CXR dataset \cite{nguyen2022vindr}, which is designed for disease classification and localization in chest radiographs. This work makes the following contributions:

\begin{itemize}    
    \item By combining advanced vision and language models, a new deep-learning framework was introduced to jointly detect diseases and localize regions of interest in chest X-ray images.
    \item Well-crafted prompts and transfer learning techniques were incorporated to enhance the framework’s performance and extend its applicability.
    \item Extensive experiments were conducted on a public dataset showcasing the success of the proposed method for chest X-ray-based disease detection and localization with a strong F1 score.
\end{itemize}

The structure of this paper is organized as follows. Section \ref{sec:related_work} offers an overview of the relevant literature. This is succeeded by Section \ref{sec:methodology}, which details the proposed methodology. Following this, Section \ref{sec:experiments} describes the experimental setup and discusses the results obtained with the proposed method. Lastly, Section \ref{sec:conclusion} concludes the paper and proposes directions for future research.

\section{Related Work}
\label{sec:related_work}
In this section, previous works that have integrated vision and language models for the analysis of chest X-rays are reviewed. The focus is on examining how these approaches have utilized the strengths of both model types to analyze visual and textual data modalities to improve disease detection, diagnosis, and report generation from chest X-ray images.

Wang et al. \cite{wang2021self} used paired chest X-ray images and text reports, as well as chest X-ray images and text reports individually from multiple institutions, to pre-train a transformer-based deep learning model in a self-supervised way. The pre-trained models were then applied to various downstream tasks, including X-ray classification, retrieval, and image regeneration. Their experimental results on these tasks showed that the proposed pre-training techniques led to more generalizable models.

To extract effective representations of textual and visual data in the medical domain, Monajatipoor et al. \cite{monajatipoor2021berthop} proposed BERTHop, a transformer-based model built upon PixelHop \cite{chen2019pixelhop} and VisualBERT \cite{li2019visualbert}. BERTHop enhances the ability to capture associations between these two modalities by first encoding the image and text, then extracting potential features from both. A transformer-based model then learns the relationships between these modalities. By utilizing appropriate vision and text extractors, BERTHop is capable of identifying abnormalities and associating them with textual labels. The model achieved high values of Area Under the Curve of the Receiver Operating Characteristic curve on imbalanced disease diagnosis datasets.

Wiehe et al. \cite{wiehe2022language} investigated the adaptation of CLIP-based models \cite{radford2021learning} for classifying chest X-ray images. Because the features learned by pre-trained CLIP on general internet data do not transfer directly to chest radiography, they introduced contrastive language supervision to tailor the model to this domain. The adapted system outperformed supervised approaches on the MIMIC-CXR dataset \cite{johnson2019mimic} and generalized well to CheXpert \cite{irvin2019chexpert} and RSNA Pneumonia \cite{wu2024pneumonia}. In addition, language supervision improved explainability, enabling the multimodal model to generate images from text and allowing experts to examine what the model had learned.

Lee et al. \cite{lee2023unixgen} introduced UniXGen, a unified model designed for chest X-ray and text report generation. The model uses a vector quantization method to convert chest X-rays into discrete visual tokens, treating both the generation of X-rays and reports as sequence generation tasks. To address scenarios where specific views of X-rays are unavailable, special tokens were introduced to generate chest X-rays with the desired views. Experimental results demonstrated that the generated multi-view chest X-rays accurately captured abnormal findings present in subsequent X-rays.

Li et al. \cite{li2024prompt} proposed a method for generating structured chest X-ray text reports by leveraging a pre-trained LLM guided by specific prompts. Their approach focuses on identifying anatomical regions within chest X-rays to generate targeted sentences that highlight key visual elements, forming the foundation of a structured report based on these anatomy-focused sentences. The method further enhances the report by converting detected anatomical regions into textual prompts that convey anatomical understanding to the LLM. Additionally, clinical context prompts are used to guide the LLM to emphasize interactivity and address clinical requirements. By integrating these anatomy-focused sentences and clinical prompts, the pre-trained LLM is able to generate structured chest X-ray reports that are tailored to the identified anatomical regions and relevant clinical contexts. The approach was evaluated using language generation and clinical effectiveness metrics, showing strong performance in producing clear and clinically relevant reports.

Thawkar et al. \cite{thawakar2024xraygpt} presented XrayGPT, a conversational vision-language model for medical imaging that can analyze chest X-rays and answer open-ended questions. Leveraging recent LLM advances such as Bard and GPT-4, the system addresses the unique challenges of interpreting biomedical images within radiology. XrayGPT links a medical visual encoder (MedClip) \cite{wang2022medclip} to a fine-tuned language model (Vicuna) \cite{chiang2023vicuna} via a linear mapping, achieving strong performance on clinically oriented visual-conversation tasks. To further enhance results, the authors curated a large set of interactive, high-quality summaries from free-text radiology reports and used them to fine-tune the LLM.

Cho et al. \cite{cho2024pretraining} introduced a novel vision-language model called PLURAL, specifically designed for difference Visual Question Answering (diff-VQA) \cite{hu2023expert} in longitudinal chest X-rays. Diff-VQA is crucial in clinical practice, where radiologists often compare pairs of X-ray images taken at different times to assess disease progression and severity changes. PLURAL leverages a pre-trained vision-language model to enhance performance in this domain. The model was developed through a step-by-step process, starting with pre-training on natural images and texts, followed by further training using longitudinal chest X-ray data. This data includes pairs of X-ray images, question-answer sets, and radiologists' reports that describe changes in lung abnormalities and diseases over time. Experimental results demonstrated that PLURAL outperformed previous methods in diff-VQA for longitudinal X-rays and showed superior performance in conventional visual question answering for single X-ray images.

Alkhaldi et al. \cite{alkhaldi2024minigpt} introduced MiniGPT-Med, a vision-language model designed to serve as a general interface for radiology diagnosis across multiple imaging modalities including X-rays, CT scans, and MRIs. Built on the MiniGPT-v2 architecture \cite{chen2023minigpt} and using a frozen EVA visual encoder paired with a fine-tuned LLaMA 2 language model, MiniGPT-Med handles a wide range of tasks such as medical report generation, disease detection, and visual question answering. To support diverse capabilities, the model incorporates task-specific instruction tokens and a prompt template architecture, enabling it to adapt to both grounding and non-grounding tasks. Evaluated across multiple datasets, MiniGPT-Med achieved state-of-the-art results in medical report generation, outperforming both specialist and generalist models. Radiologist evaluations further validated its clinical relevance, with 76\% of generated reports rated as high-quality.

While prior studies have made significant strides in either disease classification or radiology report generation from chest X-rays, few have addressed the joint modeling of disease classification alongside precise localization of regions of interest. This knowledge gap limits the development of comprehensive diagnostic systems that can both detect pathologies and interpret their spatial context. To bridge this gap, this paper introduces ChestGPT, a unified vision-language model designed to perform both disease classification and spatial localization concurrently, enabling more interpretable and clinically actionable outputs from chest X-ray analysis.

\section{ChestGPT}
\label{sec:methodology}

\subsection{Background}
\label{sec:background}
LLaMA 2 \cite{touvron2023llama} is an advanced LLM developed by Meta AI, representing the next generation of the LLaMA series. LLaMA 2 builds upon the foundation laid by its predecessor, offering improved performance across various natural language processing tasks. The model is designed to process and generate human-like text, making it highly effective for tasks such as text summarization, translation, and conversational AI. LLaMA 2 is available in multiple sizes, ranging from 7 billion to 70 billion parameters, allowing for a range of applications from lightweight deployments to more demanding tasks. The model was trained on a mixture of publicly available datasets, ensuring a broad understanding of language while adhering to high ethical standards in AI research. LLaMA 2 has been open-sourced, promoting transparency and collaboration within the AI community.

The EVA ViT \cite{fang2023eva} represents a significant advancement in the field of visual representation learning, designed to improve the performance of vision-language models on complex visual tasks. EVA ViT builds upon the foundation laid by the original ViT, which introduced the concept of treating image patches as tokens in a manner similar to words in text processing. However, EVA ViT enhances this approach by scaling up the model to one billion parameters, allowing for stronger feature extraction and better transfer learning capabilities across various vision tasks. By pretraining on a vast dataset of publicly available images, EVA ViT achieves state-of-the-art results in image classification, object detection, and segmentation tasks without requiring extensive labeled data, making it a powerful tool for visual analysis in specialized domains like medical imaging. The model's architecture also facilitates integration with language models, enabling more sophisticated multimodal applications.

\subsection{Method}
\label{sec:method}
\begin{figure*}[h!]
    \centering
    \includegraphics[width=.8\textwidth]{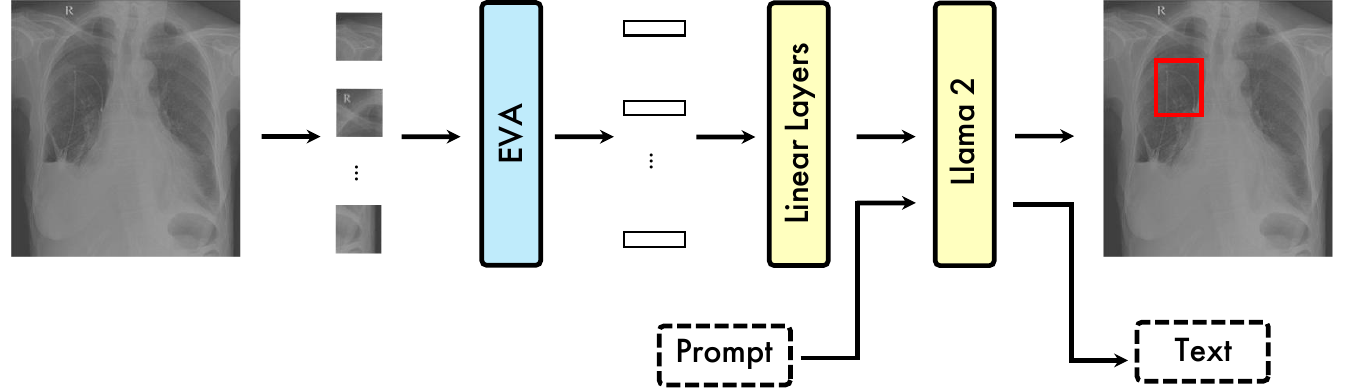}
    \caption{The block diagram of the proposed method, ChestGPT, begins with the input X-ray image, which is divided into patches. The EVA ViT then extracts feature vectors from these patches. These feature vectors are concatenated and passed through linear layers to project them into a new space, forming tokens. This projected data, along with a prompt, is fed into the LLaMA 2 model. LLaMA 2 processes this input to generate textual output that includes the detected disease and the identification of bounding boxes corresponding to the regions of interest associated with the disease. While the pre-trained EVA ViT remains frozen during training, the linear layers and LLaMA 2, which are also pre-trained, are fine-tuned to optimize their performance during the training process.}
    \label{fig:diagram}
\end{figure*}

Figure \ref{fig:diagram} illustrates the block diagram of the proposed method, ChestGPT. The input consists of a chest X-ray image along with a textual prompt, and the output is the detected and classified disease within the image, along with the location of the corresponding regions of interest.

First, the input chest X-ray image is divided into patches, which are then fed into the EVA ViT. The EVA model, originally pre-trained as described by Fang et al. \cite{fang2023eva}, remains frozen during the process, meaning its weights are not updated. EVA extracts features from these patches, which are subsequently concatenated and passed through a linear projection layer to generate tokens. These tokens, along with textual prompts, are then input into the LLaMA 2 model \cite{touvron2023llama}. LLaMA 2 processes these inputs to generate textual outputs that identify and classify diseases, as well as provide the bounding box coordinates for the regions of interest within the input image. While the EVA model remains frozen, the linear projection layers and the LLaMA 2 model are jointly trained to optimize performance in detecting and localizing diseases. The linear projection layer training process is initialized using a pre-trained model derived from MiniGPT-v2 and MiniGPT-Med \cite{chen2023minigpt, alkhaldi2024minigpt}. Both of these initializations are compared. The default prompt used for the training is as follows:

\vspace{1em}
\noindent
\textbf{[radiology] please describe this image in detail with radiological features. Use two sentences unless there are no findings. The first sentence should list the global diseases present in the image, and the second should list local diseases with localized bounding boxes.
}

\section{Experiments}
\label{sec:experiments}
In this section, the evaluation metrics and the dataset used for assessment are first explained. Following this, the experimental setup is outlined, and the results of the proposed methodology for chest X-ray image-based disease detection and classification are reported.

\subsection{Evaluation Metrics}
\label{sec:evaluation_metrics}
The performance of the proposed method is evaluated using several types of metrics, including classification performance, text validity, bounding box accuracy, and qualitative and visual analysis. For classification, accuracy and F1 score are used to assess the model’s overall effectiveness and its performance across different disease classes under varying prompts. Accuracy measures the percentage of chest X-rays for which the predicted labels exactly match the ground truth. Text validity is evaluated using Recall-Oriented Understudy for Gisting Evaluation (ROUGE-1) \cite{lin2004rouge} and Bilingual Evaluation Understudy (BLEU) \cite{papineni2002bleu} scores, which are widely used in natural language processing. Bounding box accuracy is assessed using intersection over union (IoU), which quantifies the overlap between the predicted and reference bounding boxes. IoU is calculated as the area of intersection divided by the area of the union of the two rectangles. Additionally, qualitative and visual analysis is conducted by inspecting the model’s outputs, including both the generated text and the localized image regions, to assess anomalies in chest X-rays and provide further insight into the model’s practical applicability.

\subsection{Dataset}
\label{sec:dataset}
\begin{figure*}[h!]
    \centering
    \includegraphics[width=.8\textwidth]{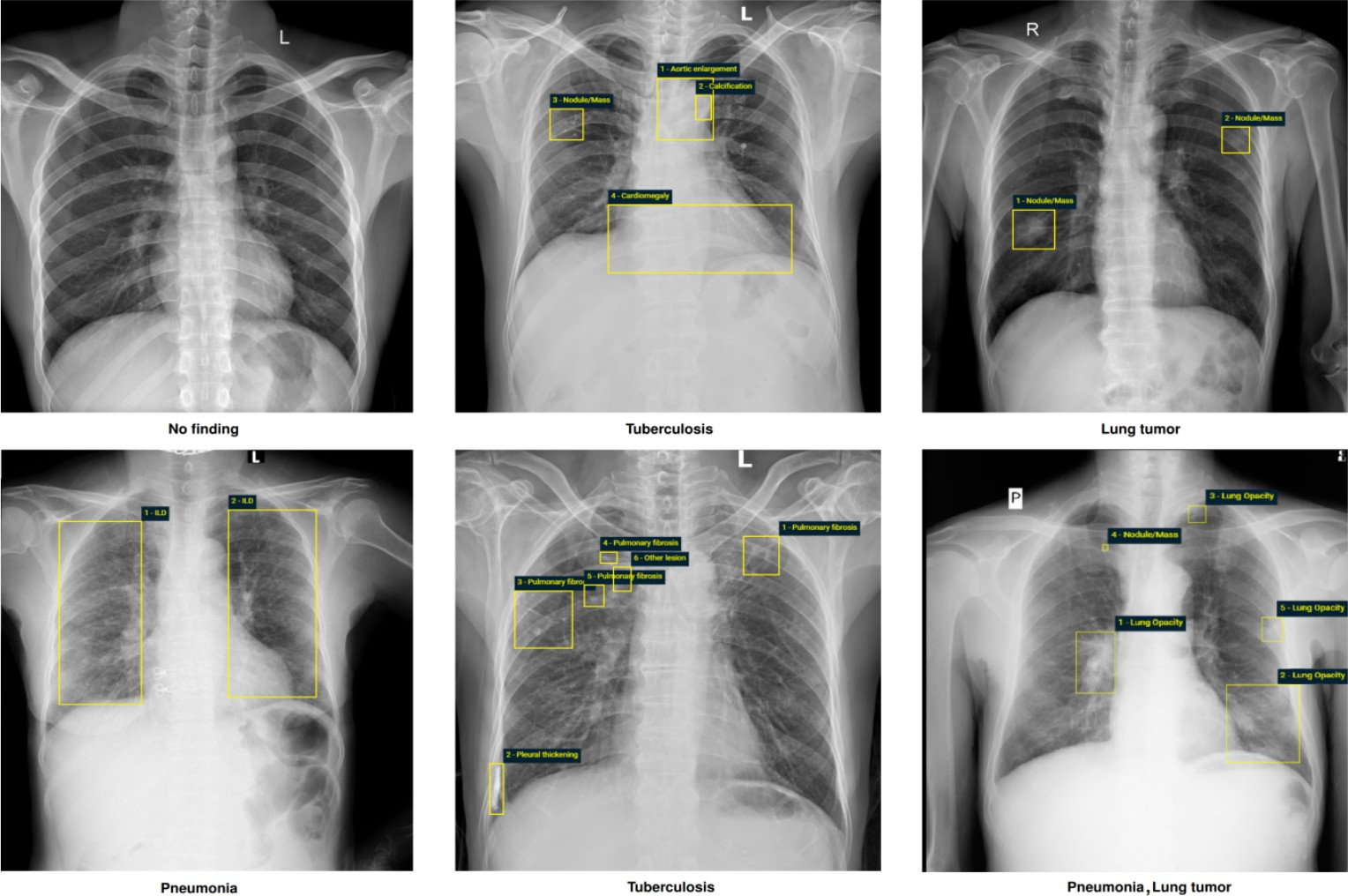}
    \caption{Example chest X-ray images from the VinDr-CXR dataset \cite{nguyen2022vindr}, showing the detected diseases and the corresponding regions of interest within the images.}
    \label{fig:dataset}
\end{figure*}

The VinDr-CXR dataset \cite{nguyen2022vindr}, which comprises posterior–anterior views of chest X-rays, was used to evaluate the proposed method. It includes a total of 18,000 images, divided into a training set of 15,000 images and a test set of 3,000 images. Each X-ray image corresponds to a different individual and is annotated with diagnoses provided independently by three radiologists. The diagnostic annotations consist of 22 local labels and 6 global labels, including a “no finding” label to indicate the absence of disease. Global labels refer to overall disease diagnoses assigned to the patient, while local labels represent specific abnormalities identified within the image. For each image, each radiologist indicated which global and local labels they would assign. In addition, bounding boxes were provided to localize each instance of a local abnormality on the X-ray. Figure \ref{fig:dataset} shows example images from the VinDr-CXR dataset, where global disease labels are listed at the bottom and local pathologies are highlighted with bounding boxes \cite{nguyen2022vindr}.

\subsection{Experimental Setup}
\label{sec:experimental_setup}
\subsubsection{Dataset Preprocessing}
\label{sec:dataset_preprocessing}
The raw image data is preprocessed by normalizing the color values to 8 bits from the original 10- to 16-bit color space in which the images were stored. This normalization is achieved using the stored windowing parameters when available; otherwise, the full-color space of the original image is used. Finally, images are resized to 448 by 448 pixels for input into the ViT. Radiology report labels are processed by converting the dataset labels into a plain text format. Two-sentence reports are generated by extracting the identified pathologies and their respective locations for each image, as stated in the prompt in subsection \ref{sec:method}. Images with conflicting labels, such as cases where one radiologist labels an image as 'no finding' while another provides a full diagnosis, are excluded from the dataset.

\subsubsection{Prompting Styles}
\label{sec:prompting_styles}
The proposed method was evaluated using three different prompting styles on the test set. The list of prompts along with their corresponding tags is presented in Table \ref{tab:prompting_styles} where the braces and the text inside them are replaced with what is said in the braces.

\aptLtoX{\begin{table*}[h!]
\caption{Sample prompts used for various prompting styles.}
\centering
\begin{tabular}{p{85pt}p{280pt}}
\hline
\textbf{Prompting Style} & \textbf{Sample Prompt} \\ \hline
\textbf{Default} & [radiology] Please describe this image in detail with radiological features. Use two sentences unless there are no findings. The first sentence should list the global diseases present in the image, and the second should list local diseases with localized bounding boxes. \\ \hline
\textbf{Global} & [radiology] please describe this chest x-ray in detail with radiological features. Given that the following abnormalities are present in the image: \{LIST OF LOCAL LABEL + BOUNDING BOX PAIRS\}. Find the global diseases present in the image. \\ \hline
\textbf{Local} & [radiology] please describe this chest x-ray in detail with radiological features. Given that the patient was diagnosed with \{LIST OF GLOBAL LABELS\}, find the local diseases/abnormalities with localized bounding boxes. \\ \hline
\end{tabular}
\label{tab:prompting_styles}
\end{table*}}{
\begin{table*}[h!]
\caption{Sample prompts used for various prompting styles.}
\centering
\begin{tabular}{>{\raggedright\arraybackslash}m{3cm} >{\justifying\noindent\arraybackslash}m{9cm}}
\hline
\textbf{Prompting Style} & \textbf{Sample Prompt} \\ \hline
\textbf{Default} & [radiology] Please describe this image in detail with radiological features. Use two sentences unless there are no findings. The first sentence should list the global diseases present in the image, and the second should list local diseases with localized bounding boxes. \\ \hline
\textbf{Global} & [radiology] please describe this chest x-ray in detail with radiological features. Given that the following abnormalities are present in the image: \{LIST OF LOCAL LABEL + BOUNDING BOX PAIRS\}. Find the global diseases present in the image. \\ \hline
\textbf{Local} & [radiology] please describe this chest x-ray in detail with radiological features. Given that the patient was diagnosed with \{LIST OF GLOBAL LABELS\}, find the local diseases/abnormalities with localized bounding boxes. \\ \hline
\end{tabular}
\label{tab:prompting_styles}
\end{table*}}

\subsubsection{Training}
\label{sec:training}
The proposed model was trained using a NVIDIA A100SXM4 GPU, an AMD EPYC 7413 CPU and 125 GB of RAM with PyTorch. Training spanned 50 epochs with a batch size of 1, 4 and 10 using the default prompt and using a batch size of 4 for the global and local prompts. The initial learning rate was set to 0.00001, with a minimum of 0.000001, a warm-up rate of 0.000001, and a weight decay of 0.05.

For 50 epochs at batch size 10, only the projection layer and LLaMA 2 were trained. LLaMA 2 fine tuning accounted for 20 hours, and the projection layer added 1 hour, for a total of 21 hours. At inference for one chest X-ray, EVA ViT forward was 0.025 seconds, the projection layer was 0.003 seconds, and LLaMA 2 decoding was 1.000 seconds, giving a total end to end latency of 1.028 seconds per image, which is acceptable for real world clinical applications.

\subsection{Experimental Results}
\label{sec:experimental_results}
Using the metrics discussed in the Methods section, the model is evaluated. First, the F1 scores and accuracies for trying to predict a diagnosis (global labels) are calculated for the MiniGPT-v2 and MiniGPT-Med initializations and shown in Table \ref{tab:global_f1}. The best results are achieved using the MiniGPT-Med initialization with the default prompt trained with a batch size of 10. The local labels are analyzed by calculating the intersection over union (IoU) of the top 1 local label, bounding box pair. Top 1 is used because it is difficult to quantify IoU when the output can have more or fewer anomalies than the ground truth. Along with the IoU, the accuracy of getting an exact match of local labels (not the bounding boxes) along with the F1 scores are also calculated and shown in Table \ref{tab:local_f1}. Lastly, Table \ref{tab:text} presents the validity of the generated text compared to the ground truth, evaluated using ROUGE-1 and BLEU scores. 

\begin{figure}[h!]
    \centering
    \subfloat[Other diseases (Test Label)]{
        \includegraphics[width=.48\linewidth]{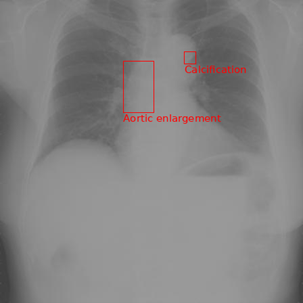}
    }\\[1ex]
    \subfloat[Other diseases (Model Output)]{
        \includegraphics[width=.47\linewidth]{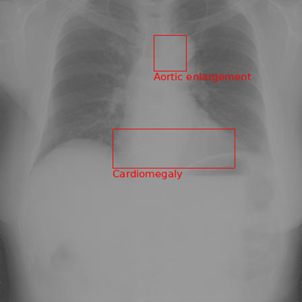}
    }
    \caption{Qualitative comparison between test labels and model output for other diseases.}
    \label{fig:qualitative_simple}
\end{figure}

As a qualitative result, Figures \ref{fig:qualitative_simple}-\ref{fig:qualitative_many_hallu} illustrates an input chest X-ray image along with the disease classification and localization generated by the proposed method using the default prompt with MiniGPT-Med initialization trained with a batch size of 10.

\begin{table*}[h!]
\caption{Aggregated models’ global label results for default prompt (batch sizes 1, 4 and 10) and global prompt. Notable results are in bold.}
\centering
\begin{tabular}{lccc}
\hline
\textbf{Prompting Style} &  \textbf{Accuracy}&\textbf{Micro F1 score}& \textbf{Macro F1 score}\\ \hline
\textbf{MiniGPT-v2:}&  &&\\ 
Default (Batch size 1)&  \textbf{0.75}&0.73& 0.27
\\ 
Default (Batch size 4)&  0.74&0.74& 0.32
\\ 
Default (Batch size 10)&  0.72&0.73& 0.31
\\ 
Global&  0.55&0.73& 0.57
\\ \hline
\textbf{MiniGPT-Med:}&  && \\ 
 Default (Batch size 1)
& 0.70& 0.69&0.20
\\
 Default (Batch size 4)
& 0.73& 0.73&0.29
\\
 Default (Batch size 10)
& \textbf{0.75}& \textbf{0.76}&0.38
\\ 
Global&  0.56&0.74& \textbf{0.60}
\\ \hline
\end{tabular}
\label{tab:global_f1}
\end{table*}

\begin{table*}[h!]
\caption{Aggregated models’ global label results for default prompt (batch sizes 1, 4 and 10) and global prompt. Notable results are in bold.}
\centering
\begin{tabular}{lcccc}
\hline
\textbf{Prompting Style} &  \textbf{Accuracy}&\textbf{Top 1 IoU}&\textbf{Micro F1 score}&\textbf{Macro F1 score}\\ \hline
\textbf{MiniGPT-v2:}&    &&&\\ 
Default (Batch size 1)&  0.04&0.11&0.38&0.15
\\ 
Default (Batch size 4)&  0.05&0.19&\textbf{0.39}&0.21
\\ 
Default (Batch size 10)&  0.03&0.21&0.38&\textbf{0.25}
\\
 Local& 0.03&\textbf{0.31}&\textbf{0.39}& 0.23
\\ \hline
\textbf{MiniGPT-Med:}&    &&&\\ 
 Default (Batch size 1)
& 0.02&0.03&0.38& 0.14
\\
 Default (Batch size 4)
& 0.03&0.09&0.37& 0.16
\\
 Default (Batch size 10)
& 0.02&0.13&0.37& 0.19
\\ 
Local&  \textbf{0.08}&0.25&\textbf{0.39}&0.23
\\ \hline
\end{tabular}
\label{tab:local_f1}
\end{table*}

\begin{table}[h!]
\caption{BLEU and ROUGE-1 scores for trained models across the different prompts and initializations. Notable results are in bold.}
\centering
\begin{tabular}{lcc}
\hline
\textbf{Prompting Style} &  \textbf{Avg. BLEU}&\textbf{Avg. ROUGE-1}\\ \hline
\textbf{MiniGPT-v2:}&  &\\ 
Default (Batch size 1)&  \textbf{0.23}&0.72\\ 
Default (Batch size 4)&  0.22&0.71\\ 
Default (Batch size 10)& \textbf{0.23}&0.71\\ 
Global&  0.06&0.41\\
 Local& 0.10& 0.33\\ \hline
\textbf{MiniGPT-Med:}&  &\\ 
 Default (Batch size 1)
& 0.22& 0.70\\
 Default (Batch size 4)
& \textbf{0.23}& 0.72\\
 Default (Batch size 10)
& \textbf{0.23}& \textbf{0.74}\\
 Global& 0.08& 0.41\\ 
Local&  0.11&0.37\\ \hline
\end{tabular}
\label{tab:text}
\end{table}

\subsection{Discussion}
\label{sec:discussion}
The model trained with the MiniGPT-Med initialization with a batch size of 10 performs the best in both accuracy and micro-averaged F1 score when predicting the global disease(s). As for the best MiniGPT-v2 model to classify global labels, the one trained with a batch size of 4 performs the best out of all MiniGPT-v2 models. It is important to note that the global prompt performs worse than the other models due to not having any 'No finding' labels. The task of finding a diagnosis using only X-rays with a diagnosis is a harder task because of the fewer data available. However, both global models outperform every other model in macro-averaged F1 score, meaning that it would perform better if all global labels were equally likely in a diagnosis. This is due to the other models possibly overfitting 'No finding' and getting high scores for correctly predicting when there is no disease in a patient. The global prompt models can be used in cases where rarer diseases are more prominent to perform better than the default prompt models.

As for local disease classification, MiniGPT-v2 performs better than its MiniGPT-Med counterpart. In general, the performance of every model when predicting local labels should be improved. As mentioned, if the local label results are low, it might be an indicator for overfitting on the “No finding” label. The local prompt has some impressive IoU numbers in comparison with the other models which indicates a better ability of pinpointing the location of abnormalities. It also has an impressive 8\% accuracy, considering that there are 22 different local labels, and each can occur multiple times, getting the exact diseases and the correct number 8\% of the time is surprising.

In the context of text validity, every model performs similarly but MiniGPT-Med trained with a batch size of 10 performs slightly better than every other model. The global and local models perform worse than the other models as they cannot score easy points by outputting 'No finding' when the patient has no disease.

Qualitatively, when looking at the example in Figure \ref{fig:qualitative_simple}, it has both the test label and the model output for diagnosis as the same, the diagnosis being 'Other Disease'. The model also does a good job of getting some of the correct local labels. There is an aortic enlargement in both the reference and model output, but just barely overlap with one another. Instead of predicting calcification, the model predicts cardiomegaly. Overall, this example comparison is a strong showing for what the model is capable of albeit on a simple (only two abnormalities) data point.

\begin{figure}[h!]
    \centering
    \subfloat[Pneumonia, Other diseases (Test Labels)]{
        \includegraphics[width=0.48\linewidth]{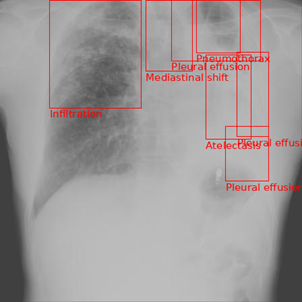}
    }\hfill
    \subfloat[Pneumonia, Tuberculosis, Other diseases (Model Outputs)]{
        \includegraphics[width=0.48\linewidth]{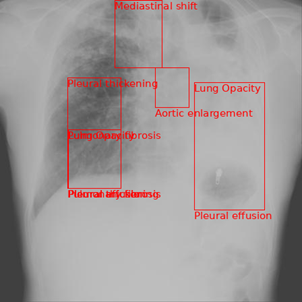}
    }
    \caption{Qualitative comparison between test labels and model output for multiple abnormalities.}
    \label{fig:qualitative_many_abnor}
\end{figure}

Figure \ref{fig:qualitative_many_abnor} shows an example with many abnormalities and multiple global labels. The model accurately predicts 'Pneumonia' and 'Other disease', but hallucinates an additional global label of 'Tuberculosis'. The model accurately predicts some of the local abnormalities with great accuracy such as the mediastinal shift at the top and pleural effusion on the right lung. Apart from this, the model does hallucinate pleural thickening and pulmonary fibrosis in the left lung and aortic enlargement while missing the infiltration and pneumothorax. Despite this, the model performs well for this more "difficult" data point.

\begin{figure}[h!]
    \centering
    \subfloat[Lung Tumor, Other diseases (Test Labels)]{
        \includegraphics[width=0.48\linewidth]{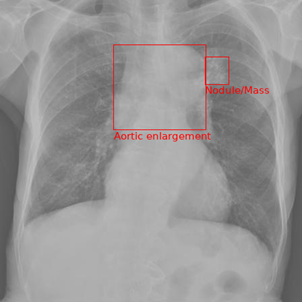}
    }\hfill
    \subfloat[Other diseases (Model Output)]{
        \includegraphics[width=0.48\linewidth]{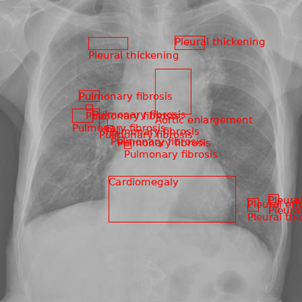}
    }
    \caption{Qualitative comparison between test labels and model output showing multiple hallucinations.}
    \label{fig:qualitative_many_hallu}
\end{figure}

Finally, Figure \ref{fig:qualitative_many_hallu} is an example of the model hallucinating abnormalities to the extreme. For what is a simple example of a lung tumor, the model outputs tons of incorrect abnormalities. When this occurs, the model should be re-run to get more accurate results as it is obvious in this case that the model output abnormalities are incorrect. However, this case could be an indicator for the radiologist that there is some sort of abnormality in the chest X-ray. Despite this, the model does accurately predict the aortic enlargement and that there is an 'Other disease' separate from the lung tumor that the model did not find.

Overall, the best performing model is the MiniGPT-Med model trained with a batch size of 10. It can classify global labels with the largest performance and its output text is most like the reference material. Depending on different situations, different models can be more helpful. For example, if the radiologist needs to find local diseases, the local prompt trained on MiniGPT-v2 can be used instead.

\section{Conclusion and Future Work}
\label{sec:conclusion}
Our research has resulted in the development of an innovative deep-learning approach for detecting and localizing diseases in chest X-ray images. In this method, the EVA ViT converts X-ray images into tokens, which are then processed by an LLM using prompts to classify and localize diseases. The successful application of our model to the VinDr-CXR dataset demonstrates its effectiveness and establishes a new paradigm for combined disease detection and localization. Additionally, the use of pre-trained weights from previous research enhances the explainability of the model by accurately detecting the bounding boxes corresponding to the identified diseases, which can assist radiologists in diagnosis and help mitigate the shortage of professionals in this field.
Future work will involve having radiologists use this model to measure how much it improves their efficiency when creating a diagnosis. To further enhance the performance of our method, future research could explore the use of more advanced vision encoders and language models, refined prompt engineering techniques, training and evaluating on more datasets, and further hyperparameter tuning with extra computation power. Additional directions include external validation across institutions, zero shot and few shot transfer, controlled benchmarks against vision only and vision language baselines under matched protocols, richer localization metrics such as mean average precision, detailed failure mode analysis by class prevalence and box size, studies of prompt robustness with practical guidance, investigations of partial and full unfreezing of the vision backbone, calibration and uncertainty estimation for clinical safety, multilingual support, and longitudinal or multiview extensions.

\begin{acks}
This research was partially supported by the Data Sciences Institute (DSI) Seed Funding for Methodologists Grant.
\end{acks}

\bibliographystyle{ACM-Reference-Format}
\bibliography{ICVGIP-Latex-Template}

\end{document}